\documentclass[conference]{IEEEtran}
\IEEEoverridecommandlockouts

\usepackage{multirow}%
\usepackage{amsmath,amssymb,amsfonts}%
\usepackage{amsthm}%
\usepackage{mathrsfs}%
\usepackage{xcolor}%
\usepackage{textcomp}%
\usepackage{manyfoot}%
\usepackage{booktabs}%
\usepackage{algorithm}%
\usepackage{algorithmicx}%
\usepackage{algpseudocode}%
\usepackage{listings}%
\usepackage{hyperref}

\usepackage{rotating}
\usepackage{caption}
\usepackage{placeins}

\definecolor{newcolor}{rgb}{.8,.349,.1}
\usepackage{colortbl}
\usepackage{graphicx}
\usepackage{caption}
\usepackage{subcaption}
\usepackage{adjustbox}

\usepackage{xspace}
\def\XS{\xspace}
\def\scu#1{\ensuremath{\mathcal{#1\XS}}}           
\def\scb#1{\ensuremath{\boldsymbol{\mathcal{#1}}}} 

\def\etal{\textit{et al.}\XS}

\def\ie{\textit{i.e.,}\XS}

   \def\Scb{{\scb{S}}\XS}

\def\Nc{{\scu{N}}\XS}

\def\gcell{\cellcolor{gray!25}}

\usepackage{amsmath}
\DeclareMathAlphabet{\mathb}{OML}{cmm}{b}{it}
\def\sbm#1{\ensuremath{\mathb{#1}}}                

\def\Cb{{\sbm{C}}\XS}  
  
  \def\eb{{\sbm{e}}\XS}

  \def\mb{{\sbm{m}}\XS}

\def\Sb{{\sbm{S}}\XS}

  \def\xb{{\sbm{x}}\XS}

\def\figureabvr{Fig.\XS}

\usepackage{cite}
\usepackage{amsmath,amssymb,amsfonts}
\usepackage{textcomp}
\usepackage{xcolor}
\def\BibTeX{{\rm B\kern-.05em{\sc i\kern-.025em b}\kern-.08em
    T\kern-.1667em\lower.7ex\hbox{E}\kern-.125emX}}
\begin{document}

\title{DiffECG: A Versatile Probabilistic Diffusion Model for ECG Signals Synthesis
\\
}
\author{\IEEEauthorblockN{Nour Neifar}
\IEEEauthorblockA{\textit{ReDCAD Lab, ENIS, University of Sfax, Tunisia} \\
nour.neifar@redcad.org}
\and
\IEEEauthorblockN{Achraf Ben-Hamadou}
\IEEEauthorblockA{\textit{SMARTS Lab, Digital Research Center of Sfax, Tunisia} \\
achraf.benhamadou@crns.rnrt.tn}
\and
\IEEEauthorblockN{Afef Mdhaffar}
\IEEEauthorblockA{\textit{ReDCAD Lab, ENIS, University of Sfax, Tunisia} \\
afef.mdhaffar@enis.tn}
\and
\IEEEauthorblockN{Mohamed Jmaiel}
\IEEEauthorblockA{\textit{ReDCAD Lab, ENIS, University of Sfax, Tunisia} \\
mohamed.jmaiel@redcad.org}
}

\maketitle

\begin{abstract}
Within cardiovascular diseases detection using deep learning applied to ECG signals, the complexities of handling physiological signals have a sparked growing interest in leveraging deep generative models for effective data augmentation. In this paper, we introduce a novel versatile approach based on denoising diffusion probabilistic models for ECG synthesis, addressing three scenarios: (i) heartbeat generation, (ii) partial signal imputation, and (iii) full heartbeat forecasting. Our approach presents the first generalized conditional approach for ECG synthesis, and our experimental results demonstrate its effectiveness for various ECG-related tasks. Moreover, we show that our approach outperforms other state-of-the-art ECG generative models and can enhance the performance of state-of-the-art classifiers.
\end{abstract}

\begin{IEEEkeywords}
Deep generative models, Diffusion model, ECG synthesis, Data augmentation, ECG forecasting
\end{IEEEkeywords}

\section{Introduction}
\label{sec:introduction}
Cardiovascular diseases (CVDs) are the leading global cause of death, emphasizing the need for heart health diagnostic tools \cite{mensah2019global}. Electrocardiograms (ECG) represent the most significant non-invasive method for identifying cardiovascular problems \cite{6518609}. ECG recordings capture the heart's electrical activity, with each heartbeat characterized by distinct waves—the P wave, QRS complex, and T wave \cite{6518609}. These waves, with their unique shapes representing specific electrical activities in the heart, provide valuable insights into the heart's rhythm, playing a crucial role in the detection of various cardiac problems. However, ECG signals present several challenges. The recording process is particularly challenging due to the imposed regulations for personal data protection and sharing \cite{monachino2023deep}. Additionally, collecting ECG data is complex due to economic constraints and time consumption \cite{monachino2023deep}. The unpredictable nature of sudden cardiac issues further complicates ECG recording, resulting in imbalanced datasets. Moreover, technical issues, including equipment failures or data transmission problems, introduce additional hurdles in ECG recording, resulting in missing data. These challenges collectively impose limitations on the effectiveness of deep learning techniques proposed for preventing CVDs. Data synthesis, imputation and forecasting using deep generative models are well-known and effective solutions for addressing these challenges. However, the synthesis of ECG signals presents a challenging task due to the complex dynamics of ECG signals \cite{golany2021ecg,nour2021Disentangling,nour022Leveraging}, which significantly vary across individual conditions and among different individuals. These complexities make it challenging to generate realistic ECG signals.


Recently, diffusion models have emerged as a highly effective class of deep generative models for these tasks \cite{ho2020denoising,song2021scorebased}. These models offer several advantages over Generative Adversarial Networks (GANs) \cite{goodfellow2014generative}, such as training stability and the ability to generate diverse synthetic samples. Diffusion models have been shown to be effective in a wide range of applications, including time series modeling \cite{alcaraz2023diffusion}.

In this context, we propose a versatile framework based on Diffusion Denoising Probabilistic Models (DDPMs) for one ECG signal generation, imputation (\ie completion), and forecasting. In contrast to other related work, our proposed method is designed to be versatile and generalized, allowing seamless adaptation across various tasks. In addition, our approach introduces a simple yet efficient conditioning encoding, allowing for an explicit transition between different synthesis tasks. Moreover, by using the spectrogram representation of ECG signals for conditioning the reverse diffusion, our method leverages insights into the frequency components of the signal. This differs from standard diffusion models for 1D signals, as it incorporates information about the frequency patterns present in ECG signals.
Our contributions are as follows: 
\begin{itemize}
\item We introduce the first versatile DDPM model for ECG signal generation, imputation, and forecasting, incorporating an efficient conditioning encoding for flexible task transitions.


\item We effectively condition the reverse diffusion based on the spectrogram representation of ECG signals to guide the ECG signal synthesis for all three tasks.


\item We provide an extensive evaluation on the MIT-BIH arrhythmia database including a comparison with the state of the art for the three different tasks, demonstrating the effectiveness of our approach.
\end{itemize}

The remainder of this paper is organized as follows: section \ref{sec:related_work} provides an overview of related work, section \ref{sec:approach} details our proposed approach, section \ref{sec:experimental_evaluation} presents the obtained experimental results, and finally, section \ref{sec:conclusion} summarizes our contributions and outlines some future research directions.

\section{Related work}\label{sec:related_work}
This section provides an overview of current research on deep generative models applied to ECG signals, as well as an introduction to the basic principles of diffusion models.
\subsection{Deep generative models applied to ECG}
Several previous studies investigated the use of deep learning techniques for time series generation and imputation, with deep generative models being a popular choice \cite{ijcai2019p429,fortuin2020gp}. GANs have been widely employed for the related ECG tasks \cite{golany2021ecg,nour2021Disentangling,nour022Leveraging,NEURIPS2018_96b9bff0}. 
For instance, the authors of \cite{nour2021Disentangling,nour022Leveraging} proposed leveraging shape prior knowledge on ECG into the generation process by using a set of anchors and 2D statistical modeling. For imputation task, GAN-based approaches were proposed. In \cite{NEURIPS2018_96b9bff0}, the authors employed a GAN framework based on a modified Gate Recurrent Unit (GRU) in the generated and discriminator networks to learn the original distribution as well as to capture the characteristics of incomplete time series. Similarly, the authors in \cite{ijcai2019p429} adopted a GRU in the denoising auto-encoder generator of the proposed GAN with the goal of an end-to-end imputation. In \cite{fortuin2020gp}, the GP-VAE framework was introduced for time series imputation, where a Gaussian process (GP) prior is used in the latent space to transform the data into a smoother and more comprehensive representation. One common challenge in GANs is mode collapse, leading to limited generation diversity. VAE models, on the other hand, struggle with learning latent variables for imputation and sampling. As stated in \cite{du2023saits}, interpreting the imputation process is challenging, since these variables could not accurately capture the specific characteristics of time series data.

Diffusion models have emerged as a successful alternative to GANs, offering improved training stability and superior generation quality, as demonstrated in \cite{alcaraz2023diffusion,alcaraz2023diffusionImputation,adib2023synthetic,Zama2023ECG}. They have found effective applications in various areas, including time series generation and imputation. Adib \etal proposed an unconditional generation method for one-channel ECG signals \cite{adib2023synthetic}. They transformed the ECG data from 1D to 2D by employing Cartesian-to-polar coordinate mapping and using techniques such as Gramian Angular Summation Fields, Gramian Angular Difference Fields, and Markov Transition Fields to create three distinct 2D embedding matrices, serving as input for a diffusion model. Unlike spectrograms, these transformations lack insight into ECG signal frequency components. Spectrograms, on the other hand, offer a comprehensive view of frequency components over time, enabling the identification of critical frequency patterns and abnormalities for accurate analysis. Furthermore, statistical metrics showed the DDPM model did not outperform a GAN-based method \cite{alcaraz2023diffusion}. 
Alcaraz et al. \cite{alcaraz2023diffusion} introduced the SSSD-ECG framework, a short 12-lead ECG generation approach based on DDPM. They employed structured state-space models (S4) as the primary component to capture long-term dependencies in time series data, in contrast to transformer layers or dilated convolutions. However, the experimental results indicated a restricted impact of synthetic ECG data on enhancing cardiac anomaly classification. The authors also introduced a method for ECG signal imputation and forecasting \cite{alcaraz2023diffusionImputation}, which is based on a conditional diffusion model combining DiffWave \cite{kong2021diffwave} and S4 models. The two distinct conditioning strategies in \cite{alcaraz2023diffusionImputation,alcaraz2023diffusion} yield a non-generalized synthesis approach, with model performance highly reliant on input type and conditional data design. Zama et al. \cite{Zama2023ECG} introduced DSAT-ECG, a novel framework for generating 12-lead ECG signals that combines a diffusion model with a State Space Augmented Transformer (SPADE). The DSAT architecture draws inspiration from SSSD-ECG, replacing S4 layers with SPADE layers. Similar to \cite{alcaraz2023diffusion}, the authors observed limited improvement in ECG signal classification tasks when using DSAT-generated synthetic ECG data.

In this paper, unlike the discussed approaches, we propose a versatile and generalized DDPM-based approach designed for easy adaptation across different tasks including ECG generation, imputation, and forecasting. In particular, we present a simple encoding of the proposed conditioning, enabling flexible and explicit transitions between distinct tasks, making it a valuable tool for scenarios where different synthesis objectives are required.

\subsection{Principle of diffusion models}
Diffusion models involve two Markovian processes: forward and reverse diffusion. In the forward process, Gaussian noise is incrementally added to the input data $\xb_0$ over $T$ steps, converting it to a standard Gaussian distribution $q(\xb_T) \sim \Nc(\xb_T; 0, I)$ using a variance schedule $\beta \in [\beta_1, \beta_T]$. In the reverse diffusion process, a neural network parameterized by $\theta$ is trained to remove the noise. The forward process is defined as : 
\begin{equation}
    q(\xb_1, .. ,\xb_t, .., \xb_T|\xb_0) = \prod_{t=1}^T q(\xb_t|\xb_{t-1})
    \label{eq:forward_process}
\end{equation}
\noindent where $q(\xb_t |\xb_{t-1})$ := $\Nc(\xb_t ; \sqrt{1 - \beta_t} \xb_{t-1} , \beta_t I)$. The closed-form expression for sampling $\xb_t$ is $\xb_t = \sqrt{\bar{\alpha_t}}\xb_0 + \sqrt{(1 - \bar{\alpha_t})}\epsilon$, where $\epsilon \sim \Nc(0, I)$, $\alpha_t = 1 - \beta_t$, and $\bar{\alpha_t} = \prod_{i=1}^t \alpha_i$. The reverse diffusion process learns to recursively denoise $\xb_t$ to retrieve $\xb_0$. It starts with pure Gaussian noise sampled from $p(\xb_T) = \Nc(\xb_T, 0, I)$, and the reverse diffusion process is described by a Markov chain as follows:
\begin{equation}
    p_{\theta}(\xb_{0:T}) = p(\xb_T) \prod_{t=1}^T p_{\theta}(\xb_{t-1}|\xb_t)
    \label{eq:reverse_process}
\end{equation}

Ho et al. \cite{ho2020denoising} showed that the reverse process can be trained with the following objective:
\begin{equation}
     L = min_\theta \text{     }E_{
   \substack{
          \!\!\!\!\!\!\!\xb_0 \sim D \\ ~~~~\epsilon\sim \Nc\! (0, I)\\
          ~~~~t \sim U(0,T)}
     }
     \lVert \epsilon - \epsilon_\theta(\sqrt{\bar {\alpha_t}}\xb_0 +\sqrt{(1-\bar {\alpha_t})}\epsilon,t) \rVert _2^2
     \label{eq:loss}
 \end{equation}
\noindent where $t$ follows the discrete uniform distribution, $D$ is the original data distribution, and the denoising function $\epsilon_\theta$ estimates the noise $\epsilon$ added to get the noisy input $\xb_t$.

\section{Proposed Approach}
\label{sec:approach}
%

\begin{figure*}
    
\centering
\includegraphics[width=\linewidth]{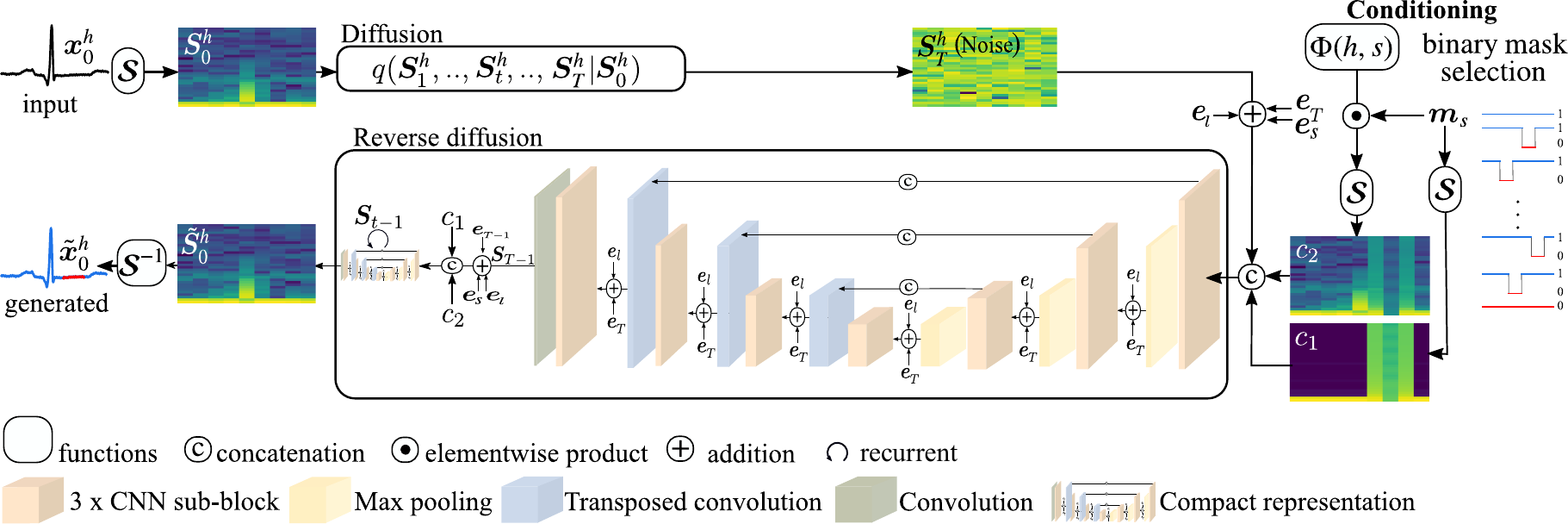}
\caption{The general principle of the proposed approach.}
\label{fig:diffusion}
\end{figure*}

The general principle of our approach is depicted in \figureabvr \ref{fig:diffusion}, which involves three main blocks: (i) ECG signal transformation, (ii) diffusion model training, and (iii) conditioning. Let us first denote $\xb_0^{h}$ as the input ECG signal, where $h$ is its heartbeat index. We use a spectrogram transformation $\Scb$ and its inverse $\Scb^{-1}$ as bidirectional transformations between 1D and 2D representations of ECG signals. Before going on the diffusion process, the input ECG signal $\xb_0^{h}$ is transformed via $\Scb(\xb_0^{h})$ to obtain $\Sb_0^{h}$. The diffusion process is then applied recursively to $\Sb_0^{h}$ by gradually adding noise to it through $T$ iterations, resulting in the noisy spectrogram $\Sb_T^{h}$. In our approach, we adapted the U-Net architecture \cite{unet_Ronneberger_2015} as the reverse diffusion model, requiring conditioning not only to specify the tasks to accomplish but also to encode the signal or part of it to be considered, depending on the selected task. Also, conditioning is crucial for task switching while maintaining a unified reverse diffusion model for all tasks; generation, imputation, or forecasting. To simplify notation, we define $\Phi (h, s)$ as the function that selects the appropriate heartbeat signal depending on $\xb_0^{h}$, and the chosen task $s$. $\Phi (h, s) = \xb_0^{h}$ if $s$ is either generation or imputation. However, if $s$ is forecasting, $\Phi (h, s) = \xb_0^{h-1}$ as forecasting requires the previous heartbeat.
Additionally, we define the binary mask $\mb_s$, a vector that explicitly indicates the values from $\Phi (h, s)$ to retain in the input ECG based on the selected task $s$. For the generation task, $\mb_{s}(i) = 0, \forall i \in [0,l]$ where $l$ is the length of the signal to generate. For signal forecasting, $\mb_{s}(i) = 1, \forall i \in [0,l]$. For the imputation task, we set $\mb_{s}(i) = 0, \forall i \in [randstart,randend]$ where $randstart$ and $randend$ represent the start and the end of the interval where the signal values are missing. 
During the reverse diffusion process, $\Sb_T^{h}$ is first concatenated to the class embedding label $\eb_l$ of $\xb_0^{h}$, the time step embedding $\eb_T$, in addition to the task-aware static embedding $\eb_s$. Moreover, we introduce two additional conditions $\Cb_1$ and $\Cb_2$. $\Cb_1$ represents the spectrogram of the mask $\mb_s$ corresponding to the chosen task $s$. $\Cb_2$ is the spectrogram of ($\xb_0^{h} \odot \mb_s$) if $s$ is the generation or imputation tasks, and the spectrogram of ($\xb_0^{h-1} \odot \mb_s$) if $s$ is the forecasting task.
The resulting concatenated tensor is then fed to our diffusion model to remove the noise added in step $T$, and generate a less noisy spectrogram ${\Sb_{T-1}^{h}}$. The model consists of seven blocks, in addition to max-pooling and deconvolution layers. Each block is comprised of three CNN sub-blocks, where each sub-block includes two convolution layers and uses the Silu activation function. This iterative process is repeated over the $T$ steps to obtain the new spectrogram $\Tilde{\Sb}^{h}_0$. 

This iterative reverse diffusion process can be formulated as follows: 
\begin{equation}
    p_{\theta}(\Sb_{0:T}^{h}|l,s,\Cb_1,\Cb_2) = p(\Sb_T^{h}) \prod_{t=1}^T p_{\theta}(\Sb_{t-1}^{h}|\Sb_t^{h},l,s,\Cb_1,\Cb_2)
    \label{eq:conditional_reverse_process}
\end{equation} 

Finally, we apply the $\Scb^{-1}$ to $\Tilde{\Sb_0^{h}}$ to produce the synthetic ECG signal $\Tilde{\xb_0^{h}}$.
To enhance the convergence of the reverse diffusion in imputation and forecasting tasks, we consider an MSE regression loss computed between generated $\Tilde{\xb_0^{h}}$ and ground truth $\xb_0^{h}$ signals. This is in addition to the reverse diffusion loss previously detailed in equation \ref{eq:loss}.

\section{Experimental evaluation}
\label{sec:experimental_evaluation}
This section presents the used dataset, details our model training settings, presents the conducted experiments, and discusses the obtained results.
\subsection{Dataset}
To train our model, we use the MIT-BIH arrhythmia database \cite{mitbih}, a widely recognized standard dataset for arrhythmia detection and classification \cite{neifar2023deep}. This dataset contains 48 half-hour ECG recordings of different patients. Each recording consists of two annotated 30-minute ECG leads, digitally recorded at 360 samples per second. This dataset includes more than 100,000 ECG heartbeats, with the majority classified as normal ECG. Three classes of heartbeats are taken into consideration in this study: the normal beats, the premature ventricular contraction beats, and fusion beats (classes N, V, and F respectively). We focused our experiments on using MLII lead since it was the most commonly employed lead to record a single ECG channel for all patients in the database and the most frequently used configuration in the previous research works.  

\subsection{Training settings}
We implemented our model using the PyTorch library and trained it on an Ubuntu server equipped with a GeForce GTX 1080 Ti GPU with 11 GB of memory. The ADAM algorithm with a learning rate of 0.001 was used for stochastic gradient optimization. The number of steps in the diffusion process was set to 1000, while the minimum of schedule noise is $\beta_0 = 0.0001$ and the maximum is $\beta_T = 0.02$. We used the Torchaudio library for transferring modalities between ECG signals and their spectrograms, in both forward and reverse directions 
\footnote{{\href{https://pytorch.org/audio/stable/transforms.html}{https://pytorch.org/audio/stable/transforms.html}}}. An ECG signal is divided into heartbeats, also known as cardiac cycles, with each heartbeat consisting of 270 voltage values. A cardiac cycle is therefore a vector with a length of 270 values ($l$ = 270), corresponding to 350 and 400 milliseconds before and after the R-peak. We randomly chose 70\% of the data for the training steps, while the remaining 30\% of the data was used for model testing. A signal-wise paradigm is adopted when splitting the database.

\subsection{Experiments and results}
Two steps of evaluation were considered to evaluate our approach: quantitative and qualitative evaluations. The quantitative evaluation involves assessing the impact of augmenting the real training dataset with our synthetic ECG on the performance of state-of-the-art arrhythmia classification baselines. This assessment includes a comparison of the performance of our generation method with other competing generation approaches. Additionally, the evaluation involves the use of various standard metrics to quantitatively assess our approach in the three considered scenarios: generation, imputation, and forecasting.

As a qualitative evaluation, we visually checked the generated ECG signals for the different scenarios to identify any inconsistencies and visual incoherence and compare them to other generated signals obtained by competing approaches.
\begin{table*}
    
\centering
\caption{Classification results for the three baselines for the different settings. Acc., Pr., Re., F1.sc. stand for Accuracy, Precision, Recall, and F1 score.}
\label{tab:classification_baseline}
\begin{tabular}{ccccccccccccc}
\cmidrule[1pt](r){2-13}
 \multicolumn{1}{c}{}     & \multicolumn{4}{c|}{\cite{acharya2017deep}}     & \multicolumn{4}{c|}{\cite{kumar2019arrhythmia}} & \multicolumn{4}{c}{\cite{kachuee2018ecg}}   \\
 \cmidrule[1pt](r){2-13}
      & \multicolumn{1}{c}{Acc.} & \multicolumn{1}{c}{Pr.} & \multicolumn{1}{c}{Re.} & F1.sc. &
      \multicolumn{1}{c}{Acc.} & \multicolumn{1}{c}{Pr.} & \multicolumn{1}{c}{Re.} & F1.sc. &
      \multicolumn{1}{c}{Acc.} & \multicolumn{1}{c}{Pr.} & \multicolumn{1}{c}{Re.} & F1.sc. \\ 
      
      \hline

Setting 1 & 0.97        & 0.93        & 0.89   & 0.91 
& 0.98        &  0.87        & 0.82   & 0.84 
&0.96        & 0.87        & 0.74   & 0.77  \\ \hline

Setting 2 & 0.98           & 0.94     & 0.91  &  0.92 
& 0.98           & 0.93     & 0.91  &  0.92  
&  0.97           & 0.87     & 0.79  &  0.82   \\ \hline

Setting 3 & 0.98           & 0.93     & 0.93  &  0.92 
& 0.98           & 0.93     & 0.93  &  0.93 
&  0.98           & 0.90     & 0.92  &  0.91   \\ \hline

Setting 4 & 0.98            & 0.95     & 0.93    & 0.94  
& 0.98        & 0.96     & 0.94    & 0.95 
& 0.99        & 0.96     & 0.95    & 0.95  \\ \hline

Setting 5 & 0.99           & 0.97     & 0.95  &  0.94 
& 0.99           & 0.97     & 0.95  &  0.96 
& 0.99           & 0.96     & 0.96  &  0.96  \\ \hline

Setting 6 & 0.99          & 0.95     & 0.93    & 0.94
& 0.99        & 0.96     & 0.95    & 0.95 
& 0.99        & 0.95     & 0.96    & 0.96  \\ \hline
\end{tabular}
%
%
\end{table*}

\subsubsection{Quantitative evaluation}
\paragraph{\underline{Classification evaluation}}
In this evaluation, we used three state-of-the-art arrhythmia classification baselines \cite{kumar2019arrhythmia,kachuee2018ecg,acharya2017deep} to assess the impact of adding synthetic data obtained by different generation methods. As a first generation baseline, we adapted the seminal GAN work \cite{goodfellow2014generative} to our context. We considered also a state-of-the-art standard GAN approach \cite{delaney2019synthesis} and two other recent GAN-based ECG generation methods \cite{nour2021Disentangling,nour022Leveraging} that successfully incorporate shape prior to guide the generation process leading to prominent results.

The arrhythmia classification baselines \cite{kumar2019arrhythmia,kachuee2018ecg,acharya2017deep} were trained following these five settings:
\begin{itemize}
    \item Setting 1: only the real training dataset is used. 

    \item Setting 2: \textbf{+} synthetic ECG signals generated by \cite{goodfellow2014generative}. 

    \item Setting 3: \textbf{+} synthetic ECG signals generated by \cite{delaney2019synthesis}.

    \item Setting 4: \textbf{+} synthetic ECG signals generated by \cite{nour2021Disentangling}.

    \item Setting 5: \textbf{+} synthetic ECG signals generated by \cite{nour022Leveraging}. 

    \item Setting 6: \textbf{+} synthetic ECG signals generated using our approach.
\end{itemize}

Table \ref{tab:classification_baseline} show the obtained classification results for the different settings. We can observe that adding synthetic ECG in the training phase systematically improves the arrhythmia classification performances of the three baselines. Moreover, our generation method outperforms the standard GANs \cite{goodfellow2014generative,delaney2019synthesis} and \cite{nour2021Disentangling} and achieves comparable results to the most advanced GAN-based approach \cite{nour022Leveraging}. It is worth noticing that these approaches are used only for data generation. However, our method is a generalized approach adapted for generation, imputation, and forecasting without any fine-tuning and reconfiguration.
\begin{table}[b]
\centering
\caption{Obtained results of the used metrics across classes (N, V, and F) for generation task.}
\label{tab:generation_metrics}

\begin{tabular}{llllllll}
\toprule
& & RMSE & MAE & FID & DTW & EMD & MMD \\
\midrule 
\multirow{4}{*}{N} 
                         & \cite{goodfellow2014generative} & 1.86e-3 & 1.47e-3 & 2.14e-2 & 29.05 & 2.35e-2 & 0.49 \\
                         & \cite{delaney2019synthesis} & 1.85e-3 & 1.42e-3 & 2.20e-2 &28.91 & 2.31e-2 & 0.56 \\
                         & \cite{nour2021Disentangling} & 1.83e-3 & 1.33e-3 & 2.10e-2 & 25.67 &2.03e-2 & 0.41 \\
                         
                         & \cite{nour022Leveraging} &  1.66e-3 & 1.25e-3  &  1.74e-2 & 24.42 &  1.84e-2 &  0.40 \\
                         &\gcell Ours & 1.68e-3 & \gcell 1.20e-3 &  \gcell  1.29e-2 &\gcell 22.10 & \gcell 1.54e-2 &  0.40 \\
\midrule
\multirow{4}{*}{V} 

                         & \cite{goodfellow2014generative} & 3.87e-3 & 2.82e-3 & 5.03e-2 & 28.10 & 4.96e-2 & 0.60 \\
                         & \cite{delaney2019synthesis} & 3.79e-3 & 2.78e-3 & 4.98e-2 & 27.62 & 4.89e-2 & 0.59 \\

                         & \cite{nour2021Disentangling} & 3.54e-3 & 2.64e-3 & 3.18e-2 & 24.04 & 3.52e-2 & 0.35 \\

                       &\cite{nour022Leveraging} &  3.02e-3 &  2.22e-3 & 1.99e-2  & 21.20 & 2.52e-2 & 0.21 \\
                           & Ours & 3.32e-3 &  2.35e-3 &  2.61e-2 & 22.03 & 2.87e-2 &  0.25 \\
\midrule
\multirow{4}{*}{F}  
  & \cite{goodfellow2014generative} & 5.37e-3 & 4.18e-3 & 8.22e-2 & 8.63 & 8.26e-2 & 0.42\\
 & \cite{delaney2019synthesis} & 5.49e-3 & 4.07e-3 & 6.18e-2 & 8.35 & 6.71e-2 & 0.44\\

   & \cite{nour2021Disentangling} & 5.16e-3 & 3.85e-3 & 4.48e-2 & 7.74 & 5.45e-2 & 0.40 \\

     & \cite{nour022Leveraging} &  4.84e-3 &  3.47e-3 &  2.96e-2 & 6.61 &  4.23e-2  &  0.25 \\
                   
                        & \gcell Ours & 4.97e-3 & \gcell  3.35e-3 &  3.97e-2 & \gcell  6.50 & \gcell  3.93e-2 & 0.26 \\
\bottomrule

\end{tabular}
\end{table}

\paragraph{\underline{Evaluation metrics}}

Following previous studies such as \cite{hazra2020synsiggan,zhu2019electrocardiogram,xu2018empirical}, we consider the Root Mean Squared Error (RMSE), Mean Absolute Error (MAE), Fréchet Inception Distance (FID), Dynamic Time Warping (DTW), Earth Mover's Distance (EMD) and Maximum Mean Discrepancy (MMD) as performance metrics.  

We first present the obtained metrics for the generation task, followed by a comparison with the metrics obtained for the imputation and forecasting tasks. 

Table \ref{tab:generation_metrics} reports the values of the used metrics for our ECG generation approach and GAN-based approaches \cite{goodfellow2014generative,delaney2019synthesis,nour2021Disentangling,nour022Leveraging} on a set of real ECG from the test set and synthetic ECG. For all metrics, lower scores mean good results. Overall, our method produces results that are competitive with the four other competing generation methods. Indeed, for all heartbeat classes and metrics, our generation method outperforms the standard GANs generation baseline. In addition, our approach outperforms the advanced GAN \cite{nour2021Disentangling} for the three classes on all metrics.
On the other hand, we achieved better results on some metrics and obtained comparable results on other metrics with the most advanced GAN \cite{nour022Leveraging}, across the 3 classes. 
This could be explained by the fact of incorporating advanced prior knowledge about ECG complex dynamics in the generation process in \cite{nour022Leveraging}. For instance, for class N, we obtained (22.10, 1.54e-2) for (DTW, EMD); while \cite{nour022Leveraging} obtained (24.42, 1.84e-2). For example, we obtained for RMSE (1.68e-3, 3.32e-3, 4.97e-3) for classes (N, V, and F) which is comparable to (1.66e-3, 3.02e-3, 4.84e-3) obtained by \cite{nour022Leveraging}. 
\begin{table}[h!]
\centering
\caption{Obtained results of the used metrics across classes (N, V, and F) for the imputation task.}
\label{tab:completion_metrics}

\begin{tabular}{llllllll}
\toprule
& & RMSE & MAE & FID & EMD & MMD \\
\midrule 
\multirow{4}{*}{N} & LSTM & 6.90e-4 & 5.59e-4 & 9.88e-3 &   1.00e-2 & 0.61 \\
                         & VAE & 8.58e-4 & 5.76e-4 & 1.40e-2 &  1.35e-2 & 0.81  \\
                         & \cite{xu2022pulseimpute} & 6.51e-4 & 4.62e-4 & 5.24e-3 & 5.90e-3 & 0.32  \\
                        &\gcell Ours &\gcell3.33e-4 & \gcell 2.21e-4& \gcell 2.28e-3  &\gcell 2.35e-3 & \gcell 0.24 \\
\midrule
\multirow{4}{*}{V} & LSTM & 3.62e-3 & 2.09e-3 & 4.00e-2 &4.47e-2 & 0.68 \\
                         & VAE & 3.51e-3 & 2.06e-3 & 5.78e-2 &  5.64e-2 & 1.02 \\
                         & \cite{xu2022pulseimpute} & 3.74e-3 & 2.28e-3 & 4.04e-2 &  4.25e-2 & 0.56 \\
                        &\gcell Ours &\gcell 1.18e-3 & \gcell 7.71e-4 & \gcell 7.13e-3  &\gcell 8.45e-3 & \gcell 0.14 \\
\midrule
\multirow{4}{*}{F}  & LSTM & 4.93e-3 & 2.80e-3 & 7.23e-2 &6.75e-2 & 0.53 \\
                         & VAE & 5.09e-2 & \gcell 2.54e-3 & 6.20e-2 &  5.11e-2 & 0.28 \\
                         & \cite{xu2022pulseimpute} & 4.86e-3 & 3.24e-3 & 3.26e-2 & \gcell 3.45e-2 & \gcell 0.12 \\
                        &\gcell Ours &\gcell 4.09e-3 & 3.53e-3 & \gcell 3.57e-2 &3.46e-2 & 0.27 \\
\bottomrule

\end{tabular}

\end{table}

\begin{table}[b]
\centering
 \caption{Obtained results of the used metrics across classes (N, V, and F) for the forecasting task.}
 \label{tab:forecasting_metrics}

\begin{tabular}{llllllll}
\toprule
& & RMSE & MAE & FID & EMD & MMD \\
\midrule 
\multirow{4}{*}{N} & LSTM & 8.57e-4 & 5.52e-4 & 7.87e-3 & 6.96e-3 & 0.30 \\
                         & VAE & 8.87e-4 & 5.93e-4 & 1.42e-2 & 1.38e-2 & 0.81 \\
                         & \cite{xu2022pulseimpute} & 9.35e-4 & 6.42e-4 & 1.47e-2 &  1.44e-2 & 0.86 \\
                        &\gcell Ours &\gcell 3.49e-4 & \gcell 2.57e-4 & \gcell 2.22e-3 &\gcell 2.65e-3 & \gcell 0.26\\
\midrule
\multirow{4}{*}{V} & LSTM & 4.03e-3 & 2.44e-3 & 5.43e-2  & 5.58e-2 & 0.87 \\
                         & VAE & 4.17e-3 & 2.36e-3 & 6.29e-2 &  6.30e-2 & 0.99  \\
                         & \cite{xu2022pulseimpute} & 3.63e-3 & 2.03e-3 & 3.38e-2  & 3.75e-2 & 0.48 \\
                        &\gcell Ours &\gcell 1.07e-3 & \gcell 6.99e-4 & \gcell 6.22e-3  &\gcell 6.92e-3 & \gcell 0.11 \\
\midrule
\multirow{4}{*}{F}  & LSTM & 5.39e-2 & 4.55e-3 & 7.33e-2  & 7.21e-2 & 0.57 \\
                         & VAE & 5.29e-2 & 3.37e-3 & 7.64e-2  & 7.06e-2 & 0.53  \\
                         & \cite{xu2022pulseimpute} &  4.78e-3 & 3.18e-3 & 3.66e-2  & 3.69e-2 & \gcell 0.14 \\
                        &\gcell Ours &\gcell 3.83e-3 & \gcell 2.77e-3 & \gcell 2.74e-2 &\gcell 3.17e-2 & 0.27  \\
\bottomrule

\end{tabular}
\end{table}
\begin{figure}[t!]
\centering
\includegraphics[width=0.98\linewidth]{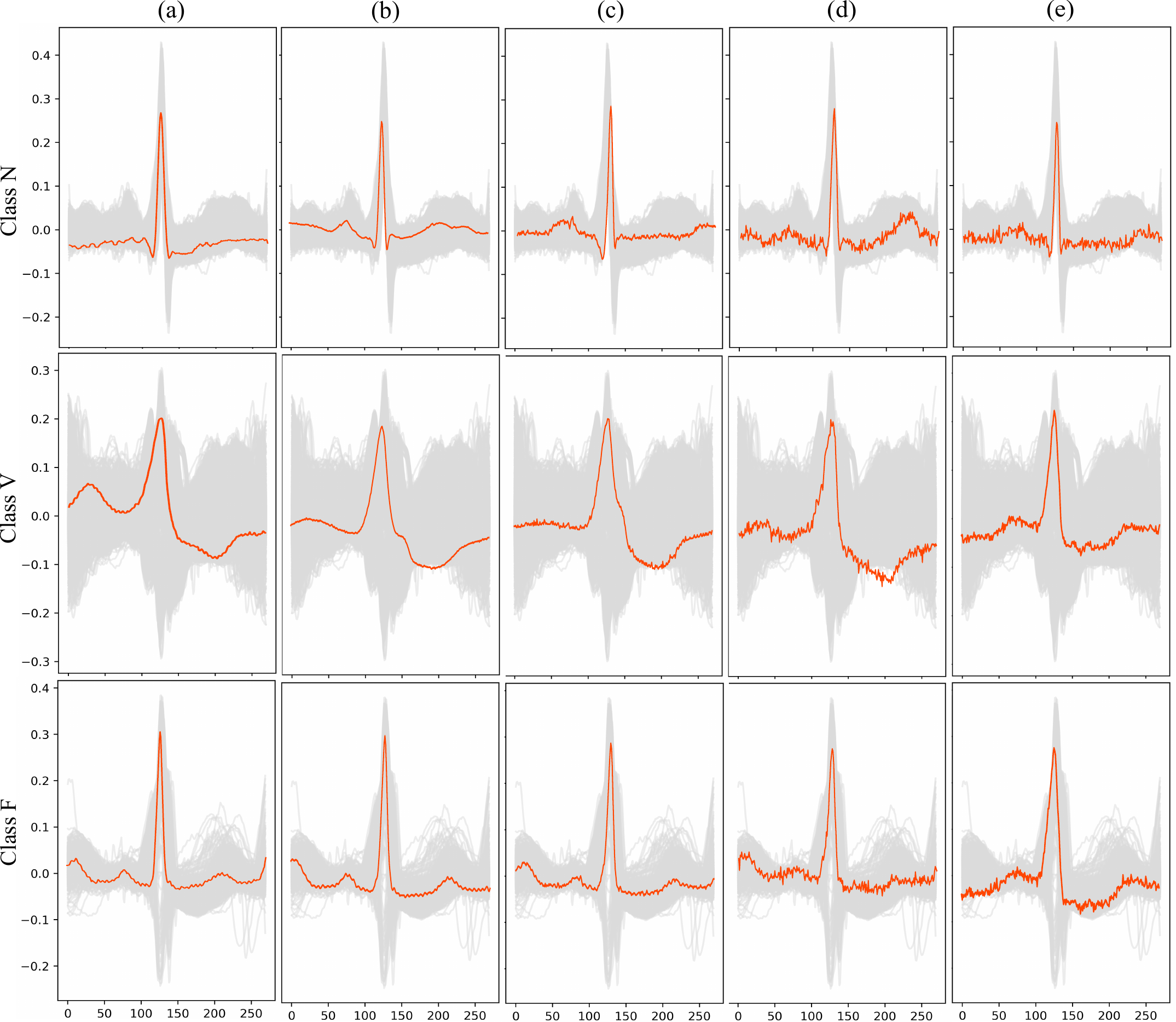}

\caption{Examples of synthetic heartbeats for classes N, V, and F classes obtained from our generation approach (a), \cite{nour022Leveraging} (b), \cite{nour2021Disentangling} (c), \cite{goodfellow2014generative} (d) and \cite{delaney2019synthesis} (e). The gray background represents the distribution of the real dataset, while the red signals depict the generated heartbeats.} 
\label{fig:fake_heartbeat}
\end{figure}
\begin{figure}[t!]
\centering
\includegraphics[width=0.98\linewidth]{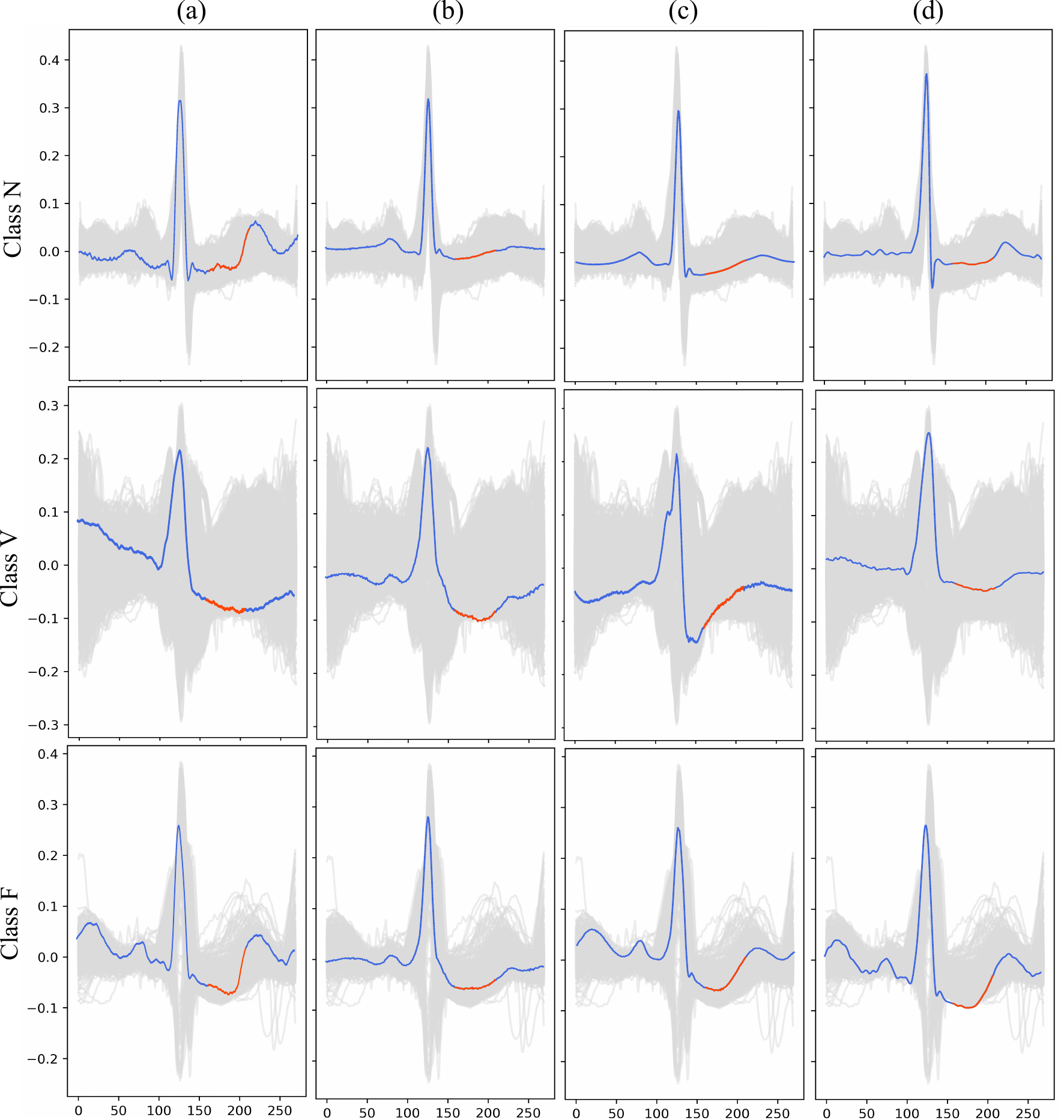}
\caption{Examples of synthetic heartbeats for classes N, V, and F classes obtained from our generation approach (a), LSTM (b), VAE (c), and \cite{xu2022pulseimpute} (d). The gray background represents the distribution of the real dataset, while the red portions depict the completed portions of the heartbeats. } 
\label{fig:fake_heartbeat_imputation}
\end{figure}
\begin{figure}[t!]
\centering
\includegraphics[width=0.98\linewidth]{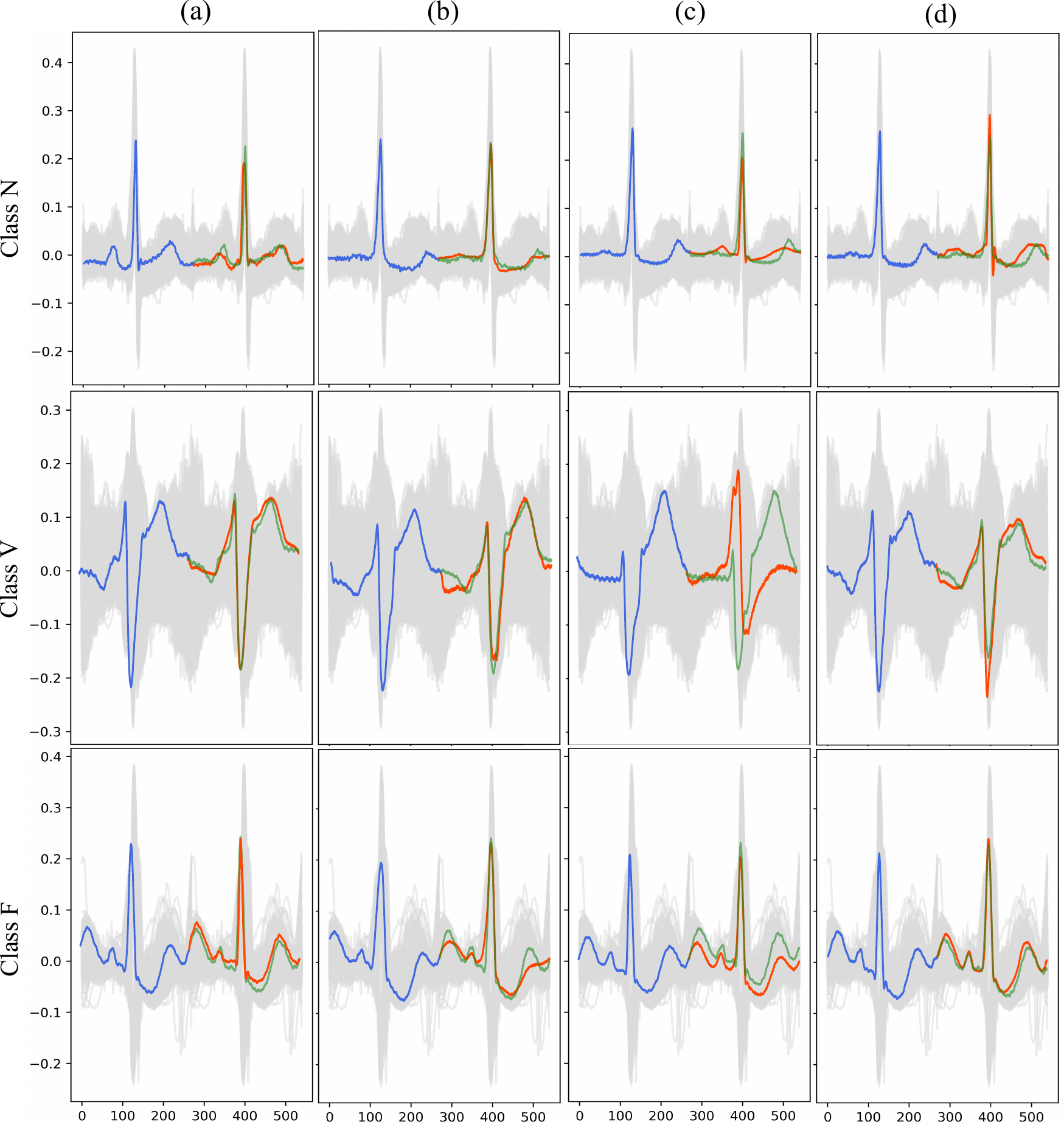}
\caption{Examples of synthetic heartbeats for classes N, V, and F classes obtained from our generation approach (a), LSTM (b), VAE (c), and \cite{xu2022pulseimpute} (d). The gray background represents the distribution of the real dataset, while the red and green portions depict the completed and ground-truth the heartbeats. } 
\label{fig:fake_heartbeat_forecasting}
\end{figure}
For the evaluation of our approach in the imputation and forecasting tasks, we compared our method to Long short-term memory (LSTM) and Variational auto-encoder (VAE) baselines in addition to a stat-of-the-art method \cite{xu2022pulseimpute} based on bottleneck dilated convolutional (BDC) self-Attention architecture. 
Tables \ref{tab:completion_metrics} and \ref{tab:forecasting_metrics} show the obtained results of the used metrics for the imputation and forecasting tasks, respectively.
We can observe that our method achieves better results globally for all metrics in both imputation and forecasting tasks. Indeed, our approach outperformed the three selected baselines in the imputation and forecasting tasks across all the metrics in classes N and V. For example, we obtained lower FID in the imputation task (2.28e-3, 7.13e-3) in classes N and V, while ((9.88e-3, 4.00e-2), (1.40e-2,5.78e-2), (5.24e-3, 4.04e-2)) in LSTM, VAE and \cite{xu2022pulseimpute}, respectively. On the other hand, we obtained superior results on RMSE and FID metrics for class F (\ie minority class) in the imputation task, while obtaining comparable results for the other metrics. Additionally, our method outperformed the other baselines on all metrics in the forecasting task except the MMD metric. For instance, we achieved 0.27 as MMD; whereas 0.14 by \cite{xu2022pulseimpute}. These results demonstrate the effectiveness of our approach in handling missing data and making accurate forecasts for ECG signals.
\subsubsection{Qualitative evaluation}
For the qualitative assessment, we start by visually comparing the synthesized signals in the three considered scenarios. We select random heartbeats obtained by our approach and the selected baselines and compare them with the distribution of the training dataset. \figureabvr {\ref{fig:fake_heartbeat}} shows examples of synthetic heartbeats from classes (N, V, and F) obtained from our generation approach, \cite{nour022Leveraging}, \cite{nour2021Disentangling}, \cite{goodfellow2014generative} and \cite{delaney2019synthesis} alongside the real distribution of these classes. Generated heartbeats from our approach and the advanced GAN methods exhibit realistic morphology and closely follow the real distribution. The generated heartbeats from standard GANs do not exhibit complete ECG morphology and are more noisy than other synthetic beats. However, the beats generated by \cite{nour2021Disentangling} are quite noisier than those obtained from our method and \cite{nour022Leveraging}.

\begin{table}[b]
\caption{Recognition rate of both real and fake heartbeats as real obtained by the two cardiologists.}
\centering
\label{medical_validation}

\begin{tabular}{c|ccccc}
\cmidrule{3-6} 
\multicolumn{2}{c}{}                                   &   Overall        & Class N & Class V & Class F \\ \midrule
\multirow{2}{*}{Cardio. 1} & $RR_{real}$ & 90\%  &94.1\%  & 85.7\%  & 83.3\%  \\ \cmidrule{2-6} 
                               & $RR_{fake}$ & 100\%  &100\%   & 100\%   & 100\%   \\ \midrule
\multirow{2}{*}{Cardio. 2} & $RR_{real}$ &96.6\%  & 100\%   & 100\%   & 83.3\%  \\ \cmidrule{2-6} 
                               & $RR_{fake}$ & 96.6\%  &100\%   & 100\%   & 83.3\%  \\ 
\bottomrule
\end{tabular}
\end{table}

\figureabvr \ref{fig:fake_heartbeat_imputation} presents examples of heartbeats with missing values in the three classes. The completed part of heartbeats by the LSTM (b) VAE (c) and \cite{xu2022pulseimpute} (d) are overly smoothed compared to completed beats by our approach (a) which is not representative of real ECG signals. Our method accurately completes the missing values in these heartbeats, demonstrating a high imputation performance. 
Additionally, \figureabvr \ref{fig:fake_heartbeat_forecasting} displays examples of heartbeats forecasting for classes (N, V, and F) using our approach and the selected baselines. The generated beats from LSTM (b), VAE (c), and \cite{xu2022pulseimpute} (d) do not always follow the ground-truth heartbeats. For example, the generated heartbeat from class V (c) does not contain the same waves as the ground-truth beat. On the other hand, \figureabvr \ref{fig:fake_heartbeat_forecasting} (a) shows our approach's ability to accurately forecast ECG signals with realistic morphology. 

The ECG signals were additionally assessed by cardiologists for a qualitative evaluation of our method. Shuffled sets containing 30 real heartbeats and 30 synthetic heartbeats from the different classes (N, V, and F) were presented to two cardiologists. The main goal is to visually differentiate between synthetic and real ECGs to assess the visual appearance of the generated ECG signals. In addition, this evaluation aimed to determine if these heartbeats could be accurately categorized into their appropriate classes, assessing the signals' class-specific properties. For this purpose, we computed the recognition rates for both real heartbeats classified as real  $RR_{real}$  and fake heartbeats classified as real $RR_{fake}$. These two metrics indicate how the generated signals are realistic and how accurately they belong to their classes. Table \ref{medical_validation} presents the obtained recognition rates by the two cardiologists. The cardiologists achieved high recognition rates for both real and synthetic heartbeats across all types. Notably, the first cardiologist classified all synthetic ECGs from the three classes as real ECGs. These results demonstrate the effectiveness of our proposed approach in generating ECG signals with realistic wave morphology.

\section{Conclusion}
\label{sec:conclusion}
In this paper, we presented the first versatile conditional diffusion framework for ECG synthesis that can perform three different tasks: heartbeats generation, imputation, and forecasting. The obtained results demonstrated the effectiveness of our approach, as well as its ability to enhance state-of-the-art classifiers’ performance. For future work, we plan to investigate the combination of diffusion models with adversarial training to further enhance ECG synthesis. Additionally, we aim to extend our approach to generate other classes of arrhythmia and multi-lead ECG signals, while also considering the synthesis of other physiological signals.

\bibliographystyle{IEEEtran}

\bibliography{arxiv}

\end{document}